\documentclass[conference]{IEEEtran}
\IEEEoverridecommandlockouts

\usepackage{times}
\usepackage{epsfig}
\usepackage{graphicx}
\usepackage{tabularx,booktabs}
\usepackage[export]{adjustbox}
\usepackage{amsmath}
\usepackage{amssymb}
\usepackage{lipsum}
\usepackage{courier}
\usepackage[font=footnotesize,subrefformat=parens,labelformat=parens]{subfig}
\usepackage{multirow}
\usepackage{threeparttable}
\usepackage{pifont}
\usepackage{enumitem}
\usepackage{tablefootnote}
\usepackage{cite}
\usepackage{url}
\usepackage{xspace}
\usepackage[T1]{fontenc}
\usepackage{dblfloatfix}
\usepackage[bottom]{footmisc}
\newcommand{\cmark}{\ding{51}}%
\newcommand{\xmark}{\ding{55}}%
\usepackage[hidelinks]{hyperref}

\makeatletter
\newcommand\footnoteref[1]{\protected@xdef\@thefnmark{\ref{#1}}\@footnotemark}
\DeclareRobustCommand\onedot{\futurelet\@let@token\@onedot}
\def\@onedot{\ifx\@let@token.\else.\null\fi\xspace}
\def\eg{\emph{e.g}\onedot} 
\def\ie{\emph{i.e}\onedot}

\def\wrt{w.r.t\onedot} 
\def\etal{\emph{et al}\onedot}
\makeatother

\def\BibTeX{{\rm B\kern-.05em{\sc i\kern-.025em b}\kern-.08em
    T\kern-.1667em\lower.7ex\hbox{E}\kern-.125emX}}

\begin{document}

\title{Image-based table recognition: data, model, and evaluation}

\author{\IEEEauthorblockN{Xu Zhong}
\IEEEauthorblockA{\textit{IBM Research Australia} \\
60 City Road, Southbank \\
 VIC 3006, Australia \\
peter.zhong@au1.ibm.com}
\and
\IEEEauthorblockN{Elaheh ShafieiBavani}
\IEEEauthorblockA{\textit{IBM Research Australia} \\
60 City Road, Southbank\\
 VIC 3006, Australia \\
elaheh.shafieibavani@ibm.com}
\and
\IEEEauthorblockN{Antonio Jimeno Yepes}
\IEEEauthorblockA{\textit{IBM Research Australia} \\
60 City Road, Southbank \\
VIC 3006, Australia \\
antonio.jimeno@au1.ibm.com}
}

\maketitle

\begin{abstract}
  Important information that relates to a specific topic in a document is often
  organized in tabular format to assist readers with information retrieval and
  comparison, which may be difficult to provide in natural language. However,
  tabular data in unstructured digital documents, \eg Portable Document Format
  (PDF) and images, are difficult to parse into structured machine-readable
  format, due to complexity and diversity in their structure and style. To
  facilitate image-based table recognition with deep learning, we develop and
  release the largest publicly available table recognition dataset
  PubTabNet\footnote{\url{https://github.com/ibm-aur-nlp/PubTabNet}}, containing 568k table images with corresponding structured
  HTML representation. PubTabNet is automatically generated by matching the XML
  and PDF representations of the scientific articles in PubMed
  Central\textsuperscript{\tiny\texttrademark} Open Access Subset (PMCOA). We
  also propose a novel attention-based encoder-dual-decoder (EDD) architecture
  that converts images of tables into HTML code. The model has a structure
  decoder which reconstructs the table structure and helps the cell decoder to
  recognize cell content. In addition, we propose a new Tree-Edit-Distance-based
  Similarity (TEDS) metric for table recognition, which more appropriately
  captures multi-hop cell misalignment and OCR errors than the pre-established
  metric. The experiments demonstrate that the EDD model can accurately
  recognize complex tables solely relying on the image representation,
  outperforming the state-of-the-art by 9.7\% absolute TEDS score.
\end{abstract}

\section{Introduction}


Information in tabular format is prevalent in all sorts of documents. Compared
to natural language, tables provide a way to summarize large quantities of data
in a more compact and structured format. Tables provide as well a format to
assist readers with finding and comparing information. An example of the
relevance of tabular information in the biomedical domain is in the curation of
genetic databases in which just between 2\% to 8\% of the information was
available in the narrative part of the article compared to the information
available in tables or files in tabular format~\cite{jimeno2014literature}.

Tables in documents are typically formatted for human understanding, and humans
are generally adept at parsing table structure, identifying table headers, and
interpreting relations between table cells. However, it is challenging for a
machine to understand tabular data in unstructured formats (\eg PDF, images) due
to the large variability in their layout and style. The key step of table understanding is to represent the unstructured tables in a machine-readable format, where the structure of the table and the content within each cell are encoded according to a pre-defined standard. This is often referred as \emph{table recognition}~\cite{gobel2013icdar}.

\begin{figure*}[!ht]
\begin{center}
  \includegraphics[width=.7\linewidth]{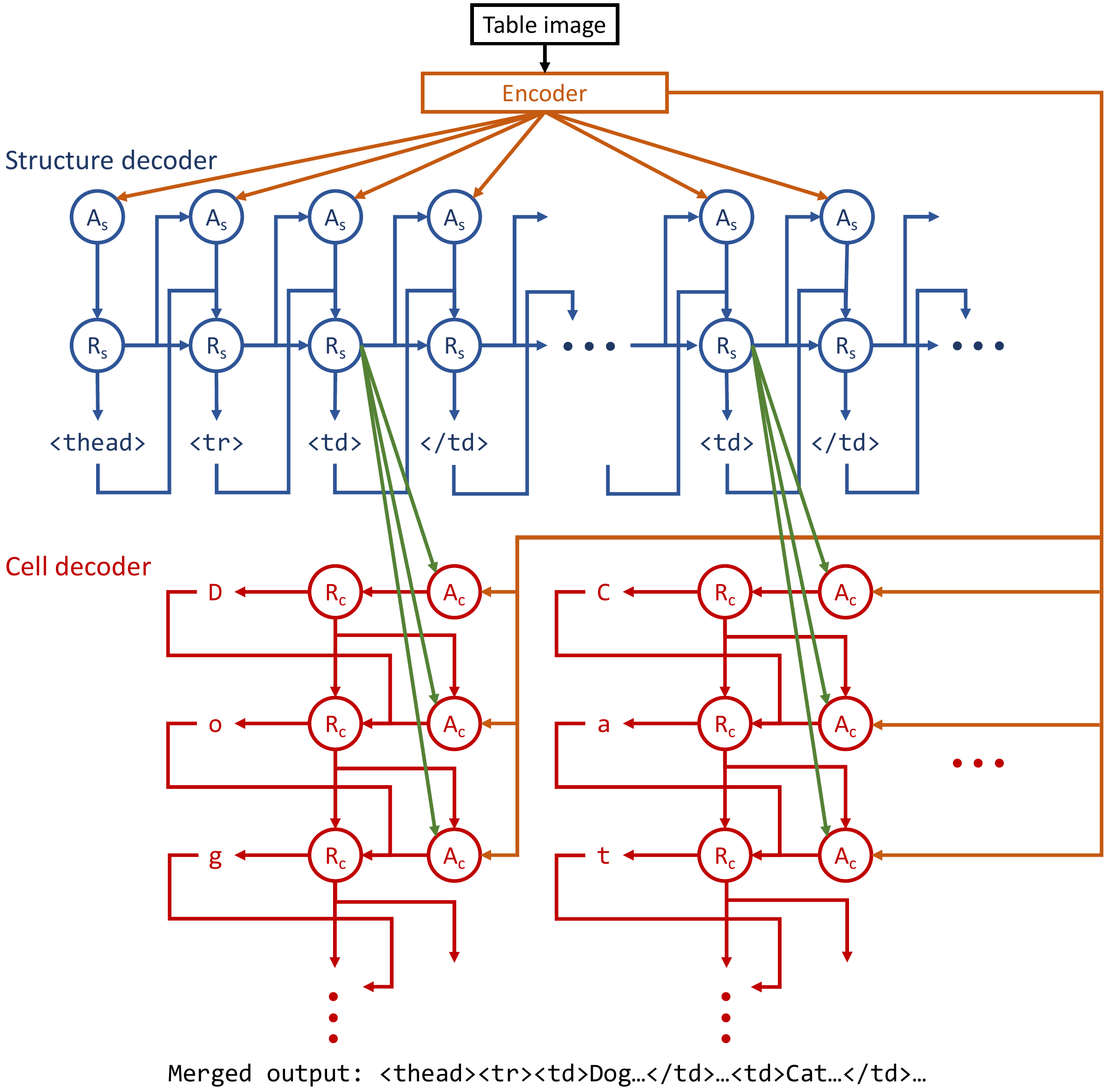}
\end{center}
  \caption{EDD architecture. The encoder is a convolutional neural network which
  captures the visual features of the input table image. $A_s$ and $A_c$
  are attention network for the structure decoder and cell decoder,
  respectively. $R_s$ and $R_c$ are recurrent units for the structure decoder
  and cell decoder, respectively. The structure decoder reconstructs table
  structure and helps the cell decoder to generate cell content. The output of
  the structure decoder and the cell decoder is merged to obtain the HTML
  representation of the input table image.}
\label{fig:arch}
\end{figure*}

This paper solves the following three problems in image-based table recognition,
where the structured representations of tables are reconstructed solely from
image input:

\setlist{nolistsep}
\begin{itemize}[noitemsep,leftmargin=*]
  \item \textbf{Data} We provide a large-scale dataset PubTabNet, which
  consists of over 568k images of heterogeneous tables extracted from the
  scientific articles (in PDF format) contained in PMCOA. By matching the metadata
  of the PDFs with the associated structured representation (provide by PMCOA\footnote{\url{https://www.ncbi.nlm.nih.gov/pmc/tools/openftlist/}} in
  XML format), we automatically annotate each table image with information about
  both the structure of the table and the text within each cell (in HTML format).
  \item \textbf{Model} We develop a novel attention-based encoder-dual-decoder (EDD)
  architecture (see Fig.~\ref{fig:arch}) which consists of an encoder, a
  structure decoder, and a cell decoder. The encoder captures the visual features
  of input table images. The structure decoder reconstructs table
  structure and helps the cell decoder to recognize cell content. Our EDD model is
  trained on PubTabNet and demonstrates superior performance compared to existing
  table recognition methods. The error analysis shows potential enhancements to
  the current EDD model for improved performance.
  \item \textbf{Evaluation} By modeling tables as a tree structure, we propose a
  new tree-edit-distance-based evaluate metric for image-based table
  recognition. We demonstrate that our new metric is superior to the
  metric~\cite{hurst2003constraint} commonly used in literature and
  competitions.
\end{itemize}

\section{Related work}
\label{sec: bkg}

\subsection{Data}

Analyzing tabular data in unstructured documents
focuses mainly on three problems: i) \emph{table detection}: localizing the
bounding boxes of tables in documents, ii) \emph{table structure recognition}:
parsing only the structural (row and column layout) information of tables, and
iii) \emph{table recognition}: parsing both the structural information and
content of table cells. Table~\ref{tab:dataset} compares the datasets that have
been developed to address one or more of these three problems. The PubTabNet
dataset and the EDD model we develop in this paper aim at the image-based table
recognition problem. Comparing to other existing datasets for table recognition
(\eg
SciTSR\footnote{\label{fn:SciTSR}https://github.com/Academic-Hammer/SciTSR},
Table2Latex~\cite{deng2019challenges}, and TIES~\cite{qasim2019rethinking}),
PubTabNet has three key advantages:

\begin{table}[!htb]
	\begin{center}
		\begin{tabularx}{0.72\linewidth}{lcccl}
			\toprule
			\textbf{Dataset} & \textbf{TD} & \textbf{TSR} & \textbf{TR} & \textbf{\# tables} \\
			\midrule
      Marmot~\cite{fang2012dataset} & \cmark & \xmark & \xmark & 958 \\
      PubLayNet~\cite{zhong2019publaynet} & \cmark & \xmark & \xmark & 113k \\
      DeepFigures~\cite{siegel2018extracting} & \cmark & \xmark & \xmark & 1.4m \\
      ICDAR2013~\cite{gobel2013icdar} & \cmark & \cmark & \cmark & 156 \\
      ICDAR2019~\cite{gao2019competition} & \cmark & \cmark & \xmark & 3.6k \\
			UNLV~\cite{shahab2010open} & \cmark & \cmark & \xmark & 558 \\
			\multirow{2}{*}{TableBank\tablefootnote{https://github.com/doc-analysis/TableBank}} & \multirow{2}{*}{\cmark} & \multirow{2}{*}{\cmark} & \multirow{2}{*}{\xmark} & 417k (TD) \\
      & & & & 145k (TSR) \\
      SciTSR\footnoteref{fn:SciTSR} & \xmark & \cmark & \cmark & 15k \\
      Table2Latex~\cite{deng2019challenges} & \xmark & \cmark & \cmark & 450k \\
      Synthetic data in~\cite{qasim2019rethinking} & \xmark & \cmark & \cmark & Unbounded \\
      \midrule
      PubTabNet & \xmark & \cmark & \cmark & 568k \\
			\bottomrule
		\end{tabularx}
	\end{center}
	\caption{Datasets for Table Detection (TD), Table Structure Recognition (TSR) and Table Recognition (TR).}
	\label{tab:dataset}
\end{table}

\setlist{nolistsep}
\begin{enumerate}[noitemsep,leftmargin=*]
  \item The tables are typeset by the publishers of over 6,000 journals in PMCOA, which
  offers considerably more diversity in table styles than other table datasets.
  \item Cells are categorized into headers and body cells, which is
  important when retrieving information from tables.
  \item The format of targeted output is HTML, which can be directly integrated
  into web applications. In addition, tables in HTML format are represented as a
  tree structure. This enables the new tree-edit-distance-based evaluation
  metric that we propose in Section~\ref{sec:metric}
\end{enumerate}

\subsection{Model}

Traditional table detection and recognition methods rely on pre-defined
rules~\cite{hirayama1995method,tupaj1996extracting,hu1999medium,gatos2005automatic,shafait2010table,paliwal2019tablenet}
and statistical machine
learning~\cite{kieninger1998t,cesarini2002trainable,e2009learning,kasar2013learning,fan2015table}.
Recently, deep learning exhibit great performance in image-based table detection
and structure recognition. Hao \etal used a set of primitive rules to propose
candidate table regions and a convolutional neural network to determine whether
the regions contain a table~\cite{hao2016table}. Fully-convolutional neural
networks, followed by a conditional random field, have also been used for table
detection~\cite{he2017multi,kavasidis2019saliency,tensmeyer2019deep}. In
addition, deep neural networks for object detection, such as
Faster-RCNN~\cite{ren2015faster}, Mask-RCNN~\cite{he2017mask}, and
YOLO~\cite{redmon2016you} have been exploited for table detection and row/column
segmentation~\cite{schreiber2017deepdesrt,gilani2017table,staar2018corpus,zhong2019publaynet}.
Furthermore, graph neural networks are used for table detection and recognition
by encoding document images as graphs~\cite{qasim2019rethinking,riba2019table}.

There are several tools (see Table~\ref{tab:test}) that can convert tables in
text-based PDF format into structured representations. However, there is limited
work on image-based table recognition. Attention-based encoder-decoder was first
proposed by Xu \etal for image captioning~\cite{xu2015show}. Deng \etal extended
it by adding a recurrent layer in the encoder for capturing long horizontal
spatial dependencies to convert images of mathematical formulas into
\LaTeX~representation~\cite{deng2017image}. The same model was trained on the
Table2Latex~\cite{deng2019challenges} dataset to convert table images into
\LaTeX~representation. As show in~\cite{deng2019challenges} and in our
experimental results (see Table~\ref{tab:test}), the efficacy of this model on
image-based table recognition is mediocre.

This paper considerably improves the performance of the attention-based
encoder-decoder method on image-based table recognition with a novel EDD
architecture. Our model differs from other existing EDD
architectures~\cite{zhou2019branchgan,morais2019learning}, where the dual
decoders are independent from each other. In our model, the cell decoder is
triggered only when the structure decoder generates a new cell. In the
meanwhile, the hidden state of the structure decoder is sent to the cell decoder
to help it place its attention on the corresponding cell in the table image.

\subsection{Evaluation}

The evaluation metric proposed in~\cite{hurst2003constraint} is commonly used in
table recognition literature and competitions. This metric first flattens the
ground truth and recognition result of a table are into a list of pairwise
adjacency relations between non-empty cells. Then precision, recall, and
F1-score can be computed by comparing the lists. This metric is simple but has
two obvious problems: 1) as it only checks immediate adjacency relations between
non-empty cells, it cannot detect errors caused by empty cells and misalignment
of cells beyond immediate neighbors; 2) as it checks relations by exact
match\footnote{Both cells are identical and the direction matches}, it does not
have a mechanism to measure fine-grained cell content recognition performance.
In order to address these two problems, we propose a new evaluation metric:
\textbf{T}ree-\textbf{E}dit-\textbf{D}istance-based \textbf{S}imilarity (TEDS).
TEDS solves problem 1) by examining recognition results at the global
tree-structure level, allowing it to identify all types of structural errors;
and problem 2) by computing the string-edit-distance when the tree-edit
operation is node substitution.

\section{Automatic generation of PubTabNet}

PMCOA contains over one million scientific articles in both unstructured (PDF)
and structured (XML) formats. A large table recognition dataset can be
automatically generated if the corresponding location of the table nodes in the
XML can be found in the PDF. Zhong \etal has proposed an algorithm to match the
the XML and PDF representations of the articles in PMCOA, which automatically
generated the PubLayNet dataset for document layout
analysis~\cite{zhong2019publaynet}. We use their algorithm to extract the table
regions from the PDF for the tables nodes in the XML. The table regions are
converted to images with a 72 pixels per inch (PPI) resolution. We use this low
PPI setting to relax the requirement of our model for high-resolution input
images. For each table image, the corresponding table node (in HTML format) is
extracted from the XML as the ground truth annotation.

It is observed that the algorithm generates erroneous bounding boxes for some
tables, hence we use a heuristic to automatically verify the bounding boxes. For
each annotation, the text within the bounding box is extracted from the PDF and
compared with that in the annotation. The bounding box is considered to be
correct if the cosine similarity of the term frequency-inverse document
frequency (Tf-idf) features of the two texts is greater than 90\% and the length
of the two texts differs less than 10\%. In addition, to improve the
learnability of the data, we remove rare tables which contains any cell that
spans over 10 rows or 10 columns, or any character that occurs less than 50
times in all the tables. Tables of which the annotation contains \texttt{math}
and \texttt{inline-formula} nodes are also removed, as we found they do not have
a consistent XML representation.

After filtering the table samples, we curate the HTML code of the tables to
remove unnecessary variations. First, we remove the nodes and attributes that
are not reconstructable from the table image, such as hyperlinks and definition
of acronyms. Second, table header cells are defined as \texttt{th} nodes in
some tables, but as \texttt{td} nodes in others. We unify the definition of
header cells as \texttt{td} nodes, which preserves the header identify of the
cells as they are still descendants of the \texttt{thead} node. Third, all the
attributes except `rowspan' and `colspan' in \texttt{td} nodes are stripped,
since they control the appearance of the tables in web browsers, which do not
match with the table image. These curations lead to consistent and clean HTML
code and make the data more learnable.

Finally, the samples are randomly partitioned into 60\%/20\%/20\%
training/\\development/test sets. The training set contains 548,592 samples.
As only a small proportion of tables contain spanning (multi-column or
multi-row) cells, the evaluation on the raw development and test sets would be
strongly biased towards tables without spanning cells. To better
evaluate how a model performs on complex table structures, we create more
balanced development and test sets by randomly drawing 5,000 tables with
spanning cells and 5,000 tables without spanning cells from the corresponding
raw set.

\section{Encoder-dual-decoder (EDD) model}

Fig.~\ref{fig:arch} shows the architecture of the EDD model, which consists of
an encoder, an attention-based structure decoder, and an attention-based cell
decoder. The use of two decoders is inspired by two intuitive considerations:
i) table structure recognition and cell content recognition are two
distinctively different tasks. It is not effective to solve both tasks at the
same time using a single attention-based decoder. ii) information in the
structure recognition task can be helpful for locating the cells that need to be
recognized. The encoder is a convolutional neural network (CNN) that captures
the visual features of input table images. The structure decoder and cell
decoder are recurrent neural networks (RNN) with the attention mechanism
proposed in~\cite{xu2015show}. The structure decoder only generates the HTML
tags that define the structure of the table. When the structure decoder
recognizes a new cell, the cell decoder is triggered and uses the hidden state
of the structure decoder to compute the attention for recognizing the content of
the new cell. This ensures a one-to-one match between the cells generated by the
structure decoder and the sequences generated by the cell decoder. The outputs
of the two decoders can be easily merged to get the final HTML representation of
the table.

\begin{figure}[!ht]
\begin{center}
  \includegraphics[width=\linewidth]{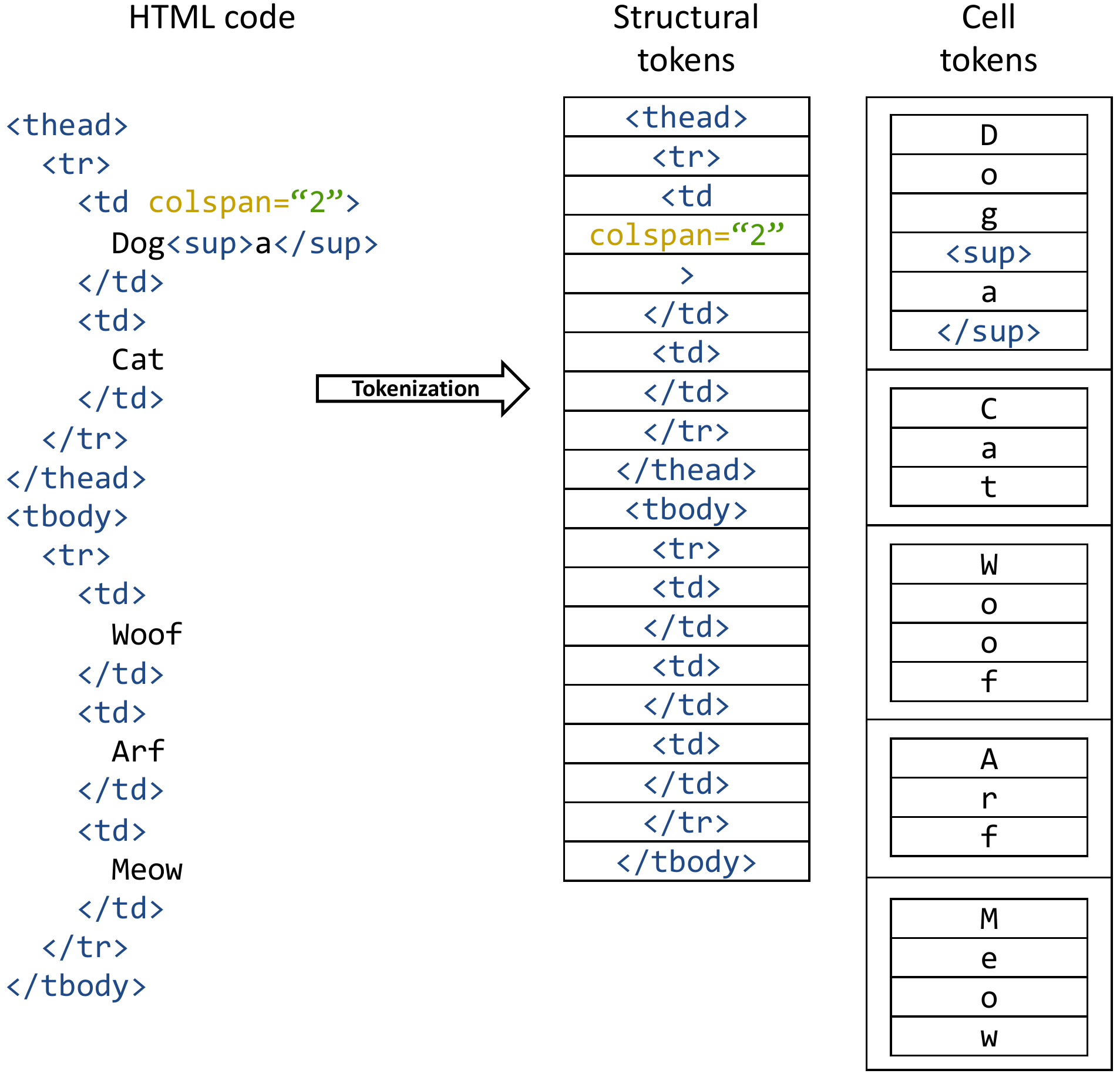}
\end{center}
  \caption{Example of tokenizing a HTML table. Structural tokens define the
  structure of the table. HTML tags in cell content are treated as single
  tokens. The rest cell content is tokenized at the character level.}
\label{fig:tokenization}
\end{figure}

As the structure and the content of an input table image are recognized
separately by two decoders, during training, the ground truth HTML
representation of the table is tokenized into structural tokens, and cell tokens
as shown in Fig.~\ref{fig:tokenization}. Structural tokens include the HTML
tags that control the structure of the table. For spanning cells, the opening
tag is broken down into multiple tokens as `\textless td', `rowspan' or
`colspan' attributes, and `\textgreater'. The content of cells is tokenized at
the character level, where HTML tags are treated as single tokens.

Two loss functions can be computed from the EDD network: i) cross-entropy loss
of generating the structural tokens ($l_s$); and ii) cross-entropy loss of
generating the cell tokens ($l_c$). The overall loss ($l$) of the EDD network is
calculated as,
\begin{equation}
  l = \lambda l_s + (1 - \lambda) l_c,
\end{equation}
where $\lambda \in [0,\, 1]$ is a hyper-parameter.

\section{Tree-edit-distance-based similarity (TEDS)}
\label{sec:metric}
Tables are presented as a tree structure in the HTML format. The root has two
children \texttt{thead} and \texttt{tbody}, which group table headers and table
body cells, respectively. The children of \texttt{thead} and \texttt{tbody}
nodes are table rows (\texttt{tr}). The leaves of the tree are table cells
(\texttt{td}). Each cell node has three attributes, \ie `colspan', `rowspan',
and `content'. We measure the similarity between two tables using the tree-edit
distance proposed by Pawlik and Augsten~\cite{pawlik2016tree}. The cost of
insertion and deletion operations is 1. When the edit is substituting a node
$n_o$ with $n_s$, the cost is 1 if either $n_o$ or $n_s$ is not \texttt{td}.
When both $n_o$ and $n_s$ are \texttt{td}, the substitution cost is 1 if the
column span or the row span of $n_o$ and $n_s$ is different. Otherwise, the
substitution cost is the normalized Levenshtein
similarity\cite{levenshtein1966binary} ($\in [0,\, 1]$) between the content of
$n_o$ and $n_s$. Finally, TEDS between two trees is computed as
\begin{equation}
TEDS(T_a,\, T_b) = 1 - \frac{EditDist(T_a,\, T_b)}{max(|T_a|,\, |T_b|)},
\end{equation}
where $EditDist$ denotes tree-edit distance, and $|T|$ is the number of nodes in
$T$. The table recognition performance of a method on a set of test samples is
defined as the mean of the TEDS score between the recognition result and ground
truth of each sample.

\begin{figure*}[!ht]
  \centering
  \begin{minipage}[b]{.4\textwidth}%
  \subfloat[Before perturbation]{\includegraphics[width=\linewidth]{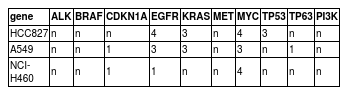}}%
  \end{minipage}%
  \begin{minipage}[b]{.4\textwidth}%
  \subfloat[After perturbation]{\includegraphics[width=\linewidth]{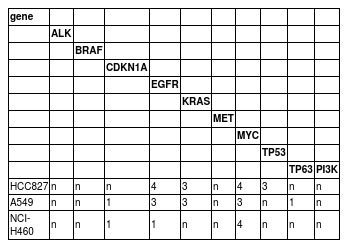}}%
  \end{minipage}%
  \caption{Example of cell shift perturbation, where 90\% of the cells in the
  first row are shifted. TEDS = 34.9\%. Adjacency relation F1 score = 80.3\%.}
  \label{fig:shift_example}
\end{figure*}

\begin{figure*}[!ht]
  \centering
  \begin{minipage}[b]{.4\textwidth}%
  \subfloat[Before perturbation]{\includegraphics[width=\linewidth]{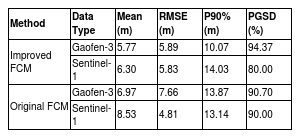}}%
  \end{minipage}%
  \begin{minipage}[b]{.4\textwidth}%
  \subfloat[After perturbation]{\includegraphics[width=\linewidth]{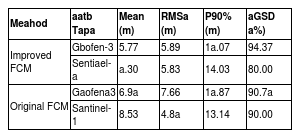}}%
  \end{minipage}%
  \caption{Example of cell content perturbationat the 10\% perturbation level. TEDS = 93.2\%. Adjacency relation F1 score = 19.1\%.}
  \label{fig:ocr_example}
\end{figure*}

In order to justify that TEDS solves the two problems of the adjacency relation
metric~\cite{hurst2003constraint} described previously in Section~\ref{sec:
bkg}, we add two types of perturbations to the validation set of PubTabNet and
examine how TEDS and the adjacency relation metric respond to the perturbations.

\setlist{nolistsep}
\begin{enumerate}[noitemsep,leftmargin=*]
  \item To demonstrate the empty-cell and multi-hop misalignment issue, we shift
  some cells in the first row downwards\footnote{If the number of rows is
  greater than the number of columns, we shift the cells in the first column
  rightwards instead.}, and pad the leftover space with empty cells. The shift
  distance of a cell is proportional to its column index. We tested 5
  perturbation levels, i.e., 10\%, 30\%, 50\%, 70\%, or 90\% of the cells in the
  first row are shifted. Fig.~\ref{fig:shift_example} shows a perturbed
  example, where 90\% of the cells in the first row are shifted.
  \item To demonstrate the fine-grained cell content recognition issue, we
  randomly modify some characters into a different one. We tested 5 perturbation
  levels, i.e., the chance that a character gets modified is set to be 10\%,
  30\%, 50\%, 70\%, or 90\%. Fig.~\ref{fig:ocr_example} shows an
  example at the 10\% perturbation level.
\end{enumerate}

\begin{figure}[!ht]
  \centering
  \begin{minipage}[b]{.5\linewidth}%
  \subfloat[Cell shift perturbation]{\includegraphics[width=\linewidth]{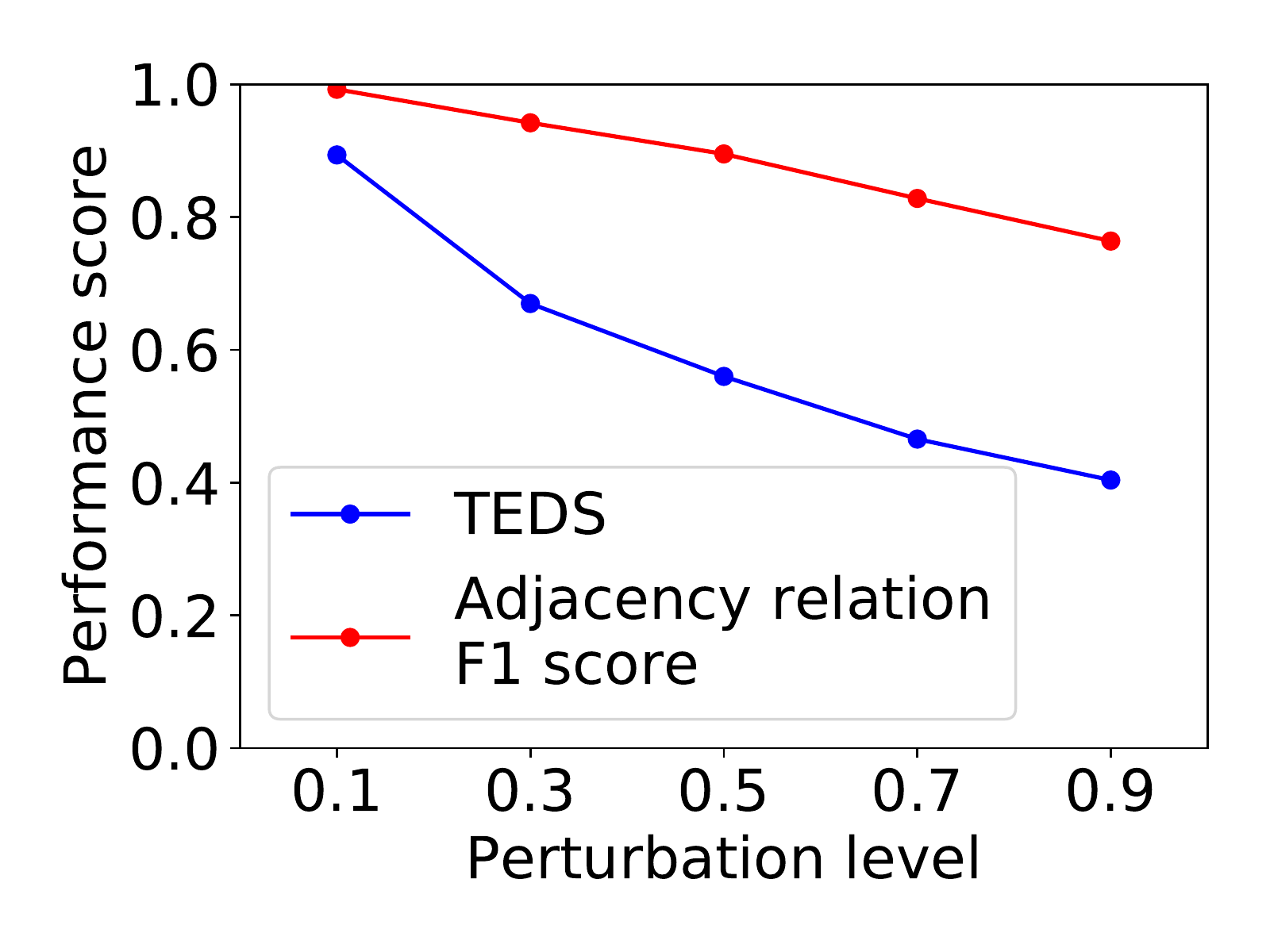}}%
  \end{minipage}%
  \begin{minipage}[b]{.5\linewidth}%
  \subfloat[Cell content perturbation]{\includegraphics[width=\linewidth]{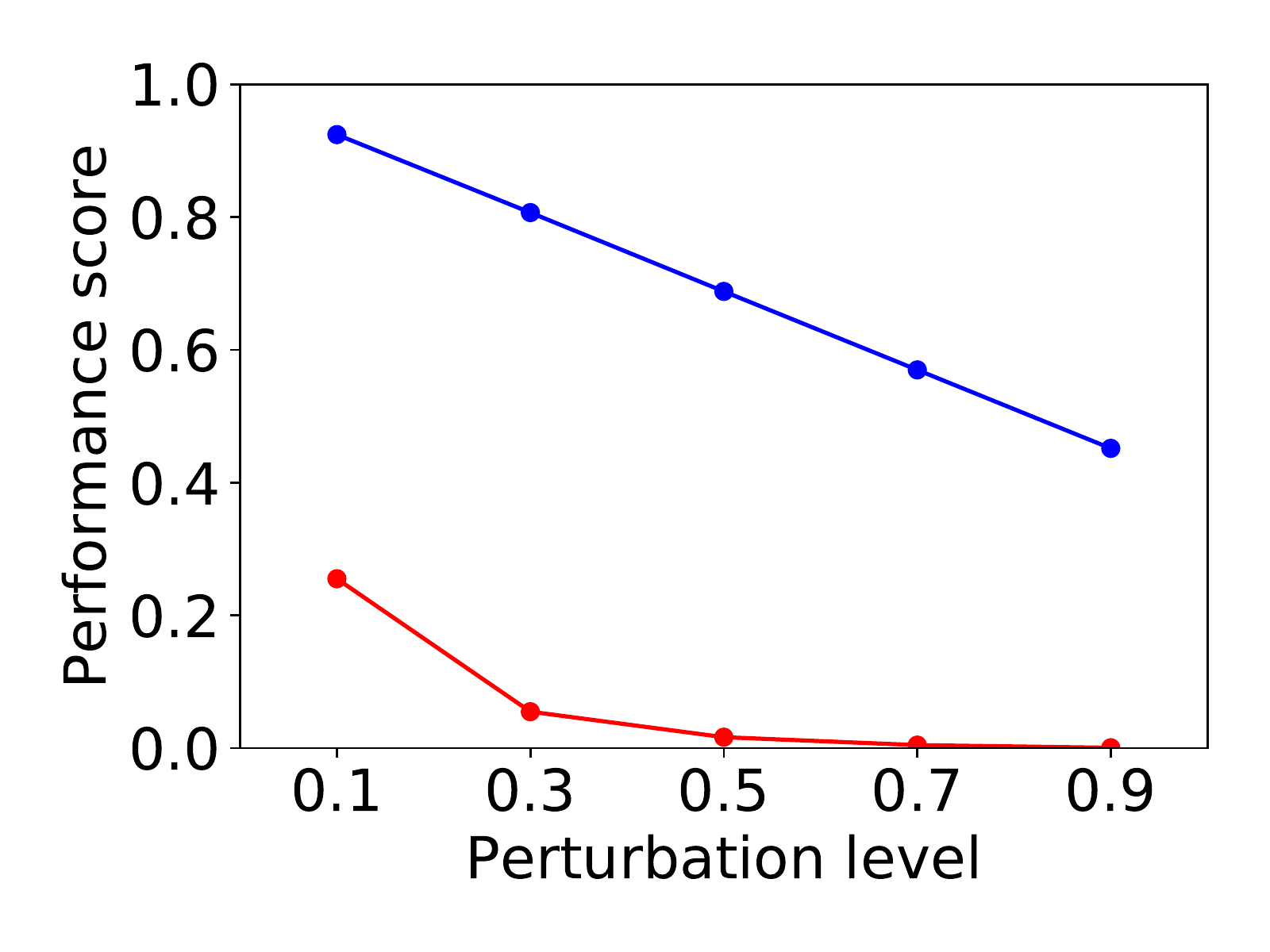}}%
  \end{minipage}%
  \caption{Comparison of the response of TEDS and the adjacency relation metric to cell shift perturbation and cell content perturbation. The adjacency relation metric is under-reacting to cell shift perturbation and over-reacting to cell content perturbation. Whereas TEDS demonstrates superiority at appropriately capturing the errors.}
  \label{fig:metric_analysis}
\end{figure}

Fig.~\ref{fig:metric_analysis} illustrates how TEDS and the adjacency relation
F1-score respond to the two types of perturbations at different levels. The
adjacency relation metric is under-reacting to the cell shift perturbation. At
the 90\% perturbation level, the table is substantially different from the
original (see example in Fig.~\ref{fig:shift_example}).
However, the adjacency relation F1-score is still nearly 80\%. On the other
hand, the perturbation causes a 60\% drop on TEDS, demonstrating that TEDS is
able to capture errors that the adjacency relation metric cannot.

When it comes to cell content perturbations, the adjacency relation metric is
over-reacting. Even the 10\% perturbation level (see example in
Fig.~\ref{fig:ocr_example}) leads to over 70\% decrease in
adjacency relation F1-score, which drops close to zero from the 50\%
perturbation level. In contrast, TEDS linearly decreases from 90\% to 40\% as
the perturbation level increases from 10\% to 90\%, demonstrating the capability
of capturing fine-grained cell content recognition errors.

\section{Experiments}

The test performance of the proposed EDD model is compared with five
off-the-shelf tools (Tabula\footnote{v1.0.4
(\url{https://github.com/tabulapdf/tabula-java})}, Traprange\footnote{v1.0
(\url{https://github.com/thoqbk/traprange})}, Camelot\footnote{v0.7.3
(\url{https://github.com/camelot-dev/camelot})},
PDFPlumber\footnote{v0.6.0-alpha (\url{https://github.com/jsvine/pdfplumber})},
and Adobe Acrobat\textsuperscript{\tiny\textregistered}
Pro\footnote{v2019.012.20040}) and the WYGIWYS model\footnote{WYGIWYS is trained
on the same samples as EDD by truncated back-propagation through time (200
steps). WYGIWYS and EDD use the same CNN in the encoder to rule out the
possibility that the performance gain of EDD is due to difference in
CNN.}~\cite{deng2017image}. We crop the test tables from the original PDF for
Tabula, Traprange, Camelot, and PDFPlumber, as they only support text-based PDF
as input. Adobe Acrobat\textsuperscript{\tiny\textregistered} Pro is tested with
both PDF tables and high-resolution table images (300 PPI). The outputs of the
off-the-shelf tools are parsed into the same tree structure as the HTML tables
to compute the TEDS score.

\subsection{Implementation details}

To avoid exceeding GPU RAM, the EDD model is trained on a subset (399k samples)
of PubTabNet training set, which satisfies
\begin{align}
  \text{width and height} & \le 512\ \text{pixels} \nonumber \\
  \text{structural tokens} & \le 300\ \text{tokens} \nonumber \\
  \text{longest cell} & \le 100\ \text{tokens}.
  \label{eq:small}
\end{align}
Note that samples in the validation and test sets are not constrained by these
criteria. The vocabulary size of the structural tokens and the cell tokens of
the training data is 32 and 281, respectively. Training images are rescaled to
$448 \times 448$ pixels to facilitate batching and each channel is normalized by
z-score.

We use the ResNet-18~\cite{he2016deep} network as the encoder. The default
ResNet-18 model downsamples the image resolution by 32. We modify the last CNN
layer of ResNet-18 to study if a higher-resolution feature map improves table
recognition performance. A total of five different settings are tested in this
paper:
\begin{itemize}
  \item EDD-S2: the default ResNet-18
  \item EDD-S1: stride of the last CNN layer set to 1
  \item EDD-S2S2: two independent last CNN layers for structure (stride=2) and cell (stride=2) decoder
  \item EDD-S2S1: two independent last CNN layers for structure (stride=2) and cell (stride=1) decoder
  \item EDD-S1S1: two independent last CNN layers for structure (stride=1) and cell (stride=1) decoder
\end{itemize}
We evaluate the performances of these five settings on the validation set and
find that a higher-resolution feature map and independent CNN layers improve
performance. As a result, the EDD-S1S1 setting provides the best validation
performance, and is therefore chosen to compare with baselines on the test set.

The structure decoder and the cell decoder are single-layer long short-term
memory (LSTM) networks, of which the hidden state size is 256 and 512,
respectively. Both of the decoders weight the feature map from the encoder with
soft-attention, which has a hidden layer of size 256. The embedding dimension of
structural tokens and cell tokens is 16 and 80, respectively. At inference time,
the output of both of the decoders are sampled with beam search (beam=3).

The EDD model is trained with the Adam~\cite{kingma2014adam} optimizer with two
stages. First, we pre-train the encoder and the structure decoder to generate
the structural tokens only ($\lambda=1$), where the batch size is 10, and the
learning rate is 0.001 in the first 10 epochs and reduced by 10 for another 3
epochs. Then we train the whole EDD network to generate both structural and cell
tokens ($\lambda=0.5$), with a batch size 8 and a learning rate 0.001 for 10
epochs and 0.0001 for another 2 epochs. Total training time is about 16 days on
two V100 GPUs.

\subsection{Quantitative analysis}

Table~\ref{tab:test} compares the test performance of the proposed EDD model and
the baselines, where the average TEDS of simple\footnote{\label{fn:simple}Tables without multi-column or multi-row cells.} and
complex\footnote{\label{fn:complex}Tables with multi-column or multi-row cells.} test tables is also shown. By solely relying on
table images, EDD substantially outperforms all the baselines on recognizing
simple and complex tables, even the ones that directly use text extracted from
PDF to fill table cells. Camelot is the best off-the-shelf tool in this
comparison. Furthermore, the performance of Adobe
Acrobat\textsuperscript{\tiny\textregistered} Pro on image input is dramatically
lower than that on PDF input, demonstrating the difficulty of recognizing tables
solely on table images. When trained on the PubTabNet dataset, WYGIWYS also
considerably outperform the off-the-shelf tools, but is outperformed by EDD by
9.7\% absolute TEDS score. The advantage of EDD to WYGIWYS is more profound on
complex tables (9.9\% absolute TEDS) than simple tables (9.5\% absolute TEDS).
This proves the great advantage of jointly training two separate decoders to
solve structure recognition and cell content recognition tasks.

\begin{table}[!hb]
  \begin{center}
    \begin{tabularx}{.75\linewidth}{llccc}
      \toprule
      \multirow{2}{*}{Input} & \multirow{2}{*}{Method} & \multicolumn{3}{c}{Average TEDS (\%)} \\
        \cline{3-5}
         & & Simple\footnoteref{fn:simple} & Complex\footnoteref{fn:complex} & All \\
      \midrule
      \multirow{5}{*}{PDF} & Tabula & 78.0 & 57.8 & 67.9 \\
       & Traprange & 60.8 & 49.9 & 55.4 \\
       & Camelot & 80.0 & 66.0 & 73.0 \\
       & PDFPlumber & 44.9 & 35.9 & 40.4 \\
       & Acrobat\textsuperscript{\tiny\textregistered} Pro & 68.9 & 61.8 & 65.3 \\
      \hline
      \multirow{3}{*}{Image} & Acrobat\textsuperscript{\tiny\textregistered} Pro & 53.8 & 53.5 & 53.7 \\
       & WYGIWYS & 81.7 & 75.5 & 78.6 \\
       & \textbf{EDD} & \textbf{91.2} & \textbf{85.4} & \textbf{88.3} \\
      \bottomrule
    \end{tabularx}
  \end{center}
\caption{Test performance of EDD and 7 baseline approaches. Our EDD model, by
solely relying on table images, substantially outperforms all the baselines.}
\label{tab:test}
\end{table}

\begin{figure*}[!ht]
  \begin{center}
  \begin{tabularx}{\textwidth}{@{\hskip 0in}c@{\hskip 0in}c@{\hskip 0in}}
    \includegraphics[valign=t,width=.49\linewidth]{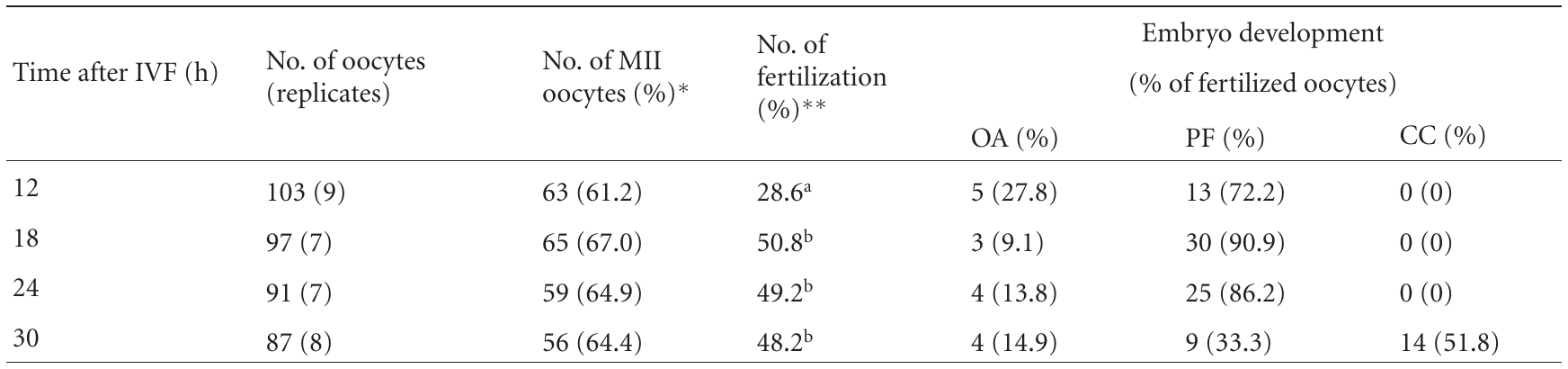} &
    \includegraphics[valign=t,trim={0 0 0 6},width=.49\linewidth]{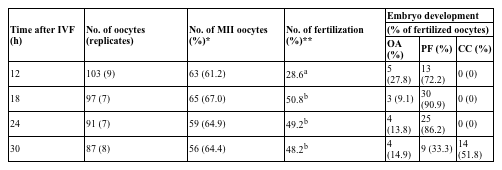} \\
    (a) Input table & (b) Ground truth \\
    \includegraphics[valign=t,trim={0 0 0 6},width=.49\linewidth]{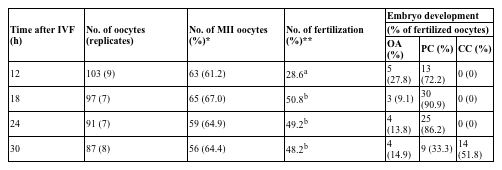} &
    \includegraphics[valign=t,trim={0 0 0 6},width=.49\linewidth]{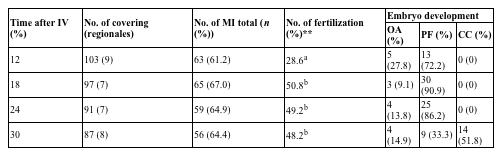} \\
    (c) EDD (TEDS $=99.8\%$) & (d) WYGIWYS (TEDS $=89.8\%$) \\
    \includegraphics[valign=t,trim={0 0 0 6},width=.49\linewidth]{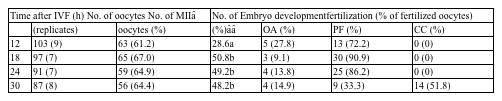} &
    \includegraphics[valign=t,trim={0 0 0 6},width=.49\linewidth]{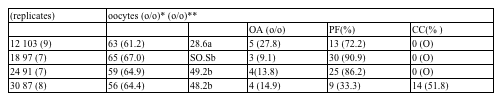} \\
    (e) Acrobat\textsuperscript{\tiny\textregistered} on PDF (TEDS $=74.8\%$) & (f) Acrobat\textsuperscript{\tiny\textregistered} on Image (TEDS $=64.2\%$) \\
    \includegraphics[valign=t,trim={0 0 0 6},width=.49\linewidth]{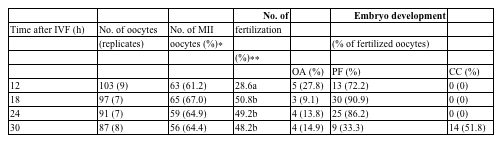} &
    \includegraphics[valign=t,trim={0 0 0 6},width=.49\linewidth]{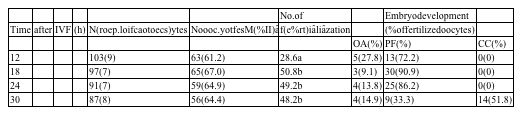} \\
    (g) Tabula (TEDS $=47.5\%$) & (h) Traprange (TEDS $=40.2\%$) \\
    \includegraphics[valign=t,trim={0 0 0 6},width=.49\linewidth]{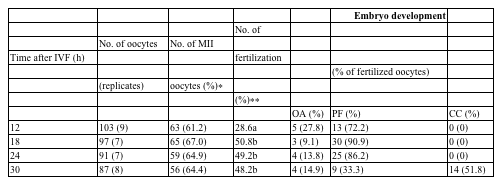} &
    \includegraphics[valign=t,trim={0 0 0 6},width=.49\linewidth]{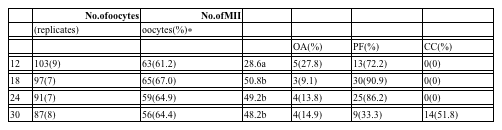} \\
    (i) Camelot (TEDS $=35.5\%$) & (j) PDFPlumber (TEDS $=30.0\%$)
  \end{tabularx}
  \end{center}
  \caption{Table recognition results of EDD and 7 baseline approaches on an
  example input table which has a complex header structure (4 multi-row (span=3)
  cells, 2 multi-column (span=3) cells, and three normal cells). Our EDD model
  perfectly recognizes the complex structure and cell content of the table,
  whereas the baselines struggle with the complex table header.}
  \label{fig:sample_results}
\end{figure*}

\begin{figure*}[!ht]
  \begin{center}
    \includegraphics[width=\linewidth]{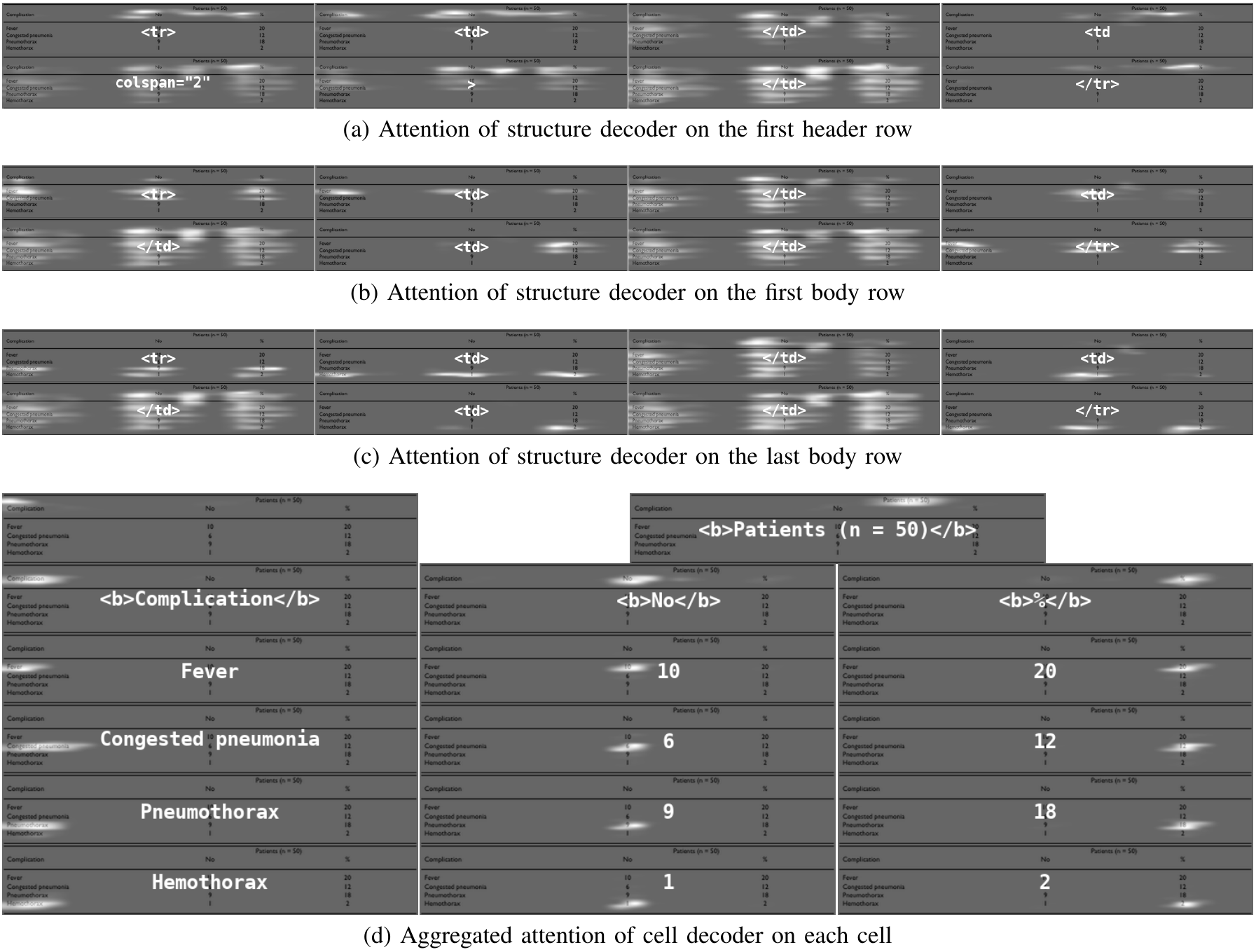}
  \end{center}
  \caption{Attention distribution of the structure decoder (a - c) and the
  cell decoder (d) on an example input table. The texts at the center of the
  images are predictions of the EDD model. The structure decoder focuses its
  attention around table cells when recognizing new rows and cells, whereas the
  cell decoder places more focused attention on cell content.}
  \label{fig:attention}
\end{figure*}

\subsection{Qualitative analysis}

To illustrate the differences in the behavior of the compared methods,
Fig.~\ref{fig:sample_results} shows the rendering of the predicted HTML given an
example input table. The table has 7 columns, 3 header rows, and 4 body rows.
The table header has a complex structure, which consists of 4 multi-row (span=3)
cells, 2 multi-column (span=3) cells, and three normal cells. Our EDD model is
able to generate an extremely close match to the ground truth, making no error
in structure recognition and a single optical character recognition (OCR) error
(`PF' recognized as `PC'). The second header row is missing in the results of
WYGIWYS, which also makes a few errors in the cell content. On the other hand,
the off-the-shelf tools make substantially more errors in recognizing the
complex structure of the table headers. This demonstrates the limited capability
of these tools on recognizing complex tables.

Figs.~\ref{fig:attention} (a) - (c) illustrate the attention of the structure
decoder when processing an example input table. When a new row is recognized
(`\textless tr\textgreater' and `\textless /tr\textgreater'), the structure
decoder focuses its attention around the cells in the row. When the opening tag
(`\textless td\textgreater') of a new cell is generated, the structure decoder
pays more attention around the cell. For the closing tag `\textless
/td\textgreater' tag, the attention of the structure decoder spreads across the
image. Since `\textless /td\textgreater' always follows the `\textless
td\textgreater' or `\textgreater' token, the structure decoder relies on the
language model rather than the encoded feature map to predict it.
Fig.~\ref{fig:attention} (d) shows the aggregated attention of the cell decoder
when generating the content of each cell. Compared to the structure decoder, the
cell decoder has more focused attention, which falls on the cell content that is
being generated.

\subsection{Error analysis}

\begin{figure}[!ht]
  \centering
  \begin{minipage}[b]{.5\linewidth}%
  \subfloat{\includegraphics[width=\linewidth]{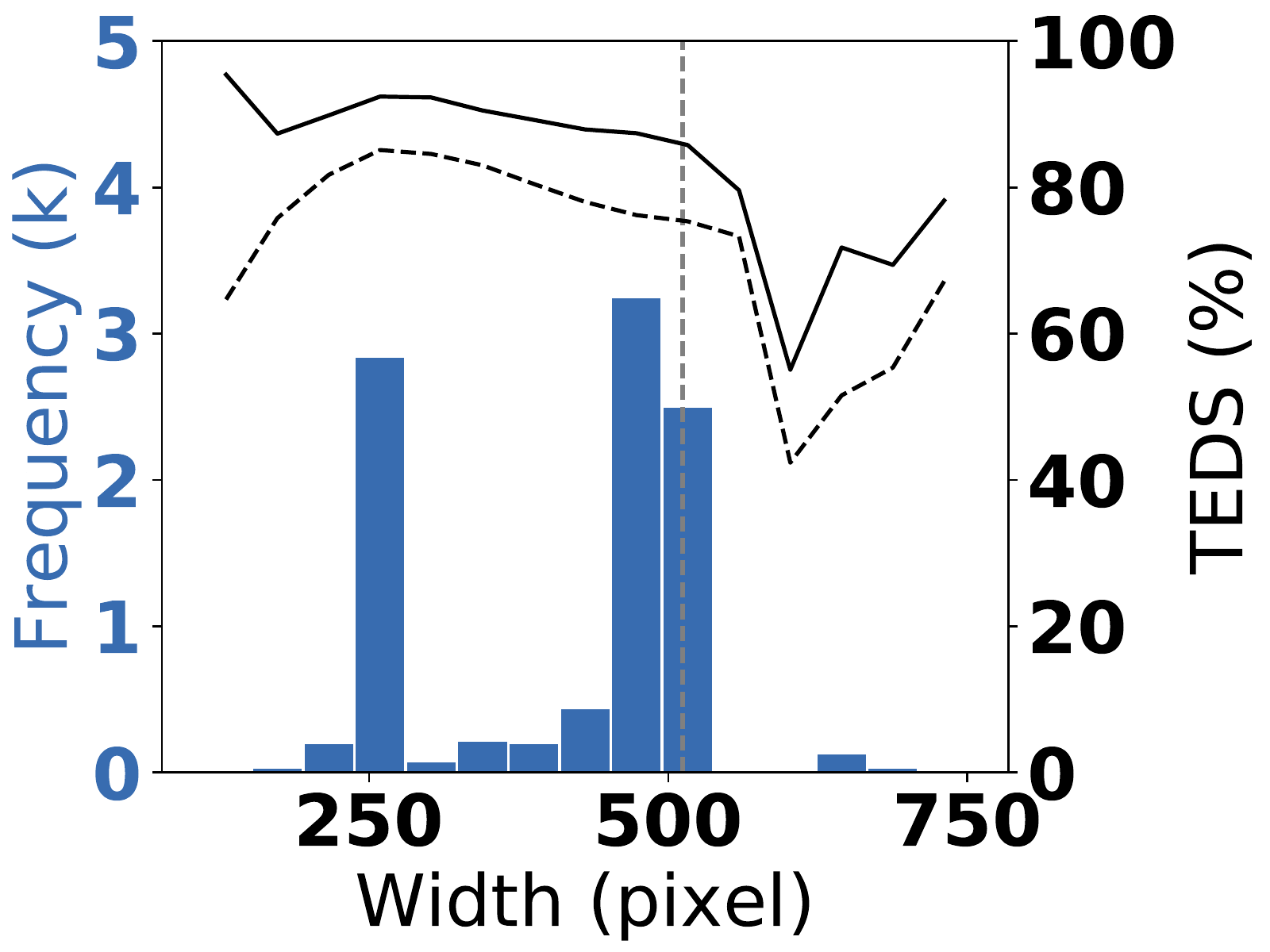}}%
  \end{minipage}%
  \begin{minipage}[b]{.5\linewidth}%
  \subfloat{\includegraphics[width=\linewidth]{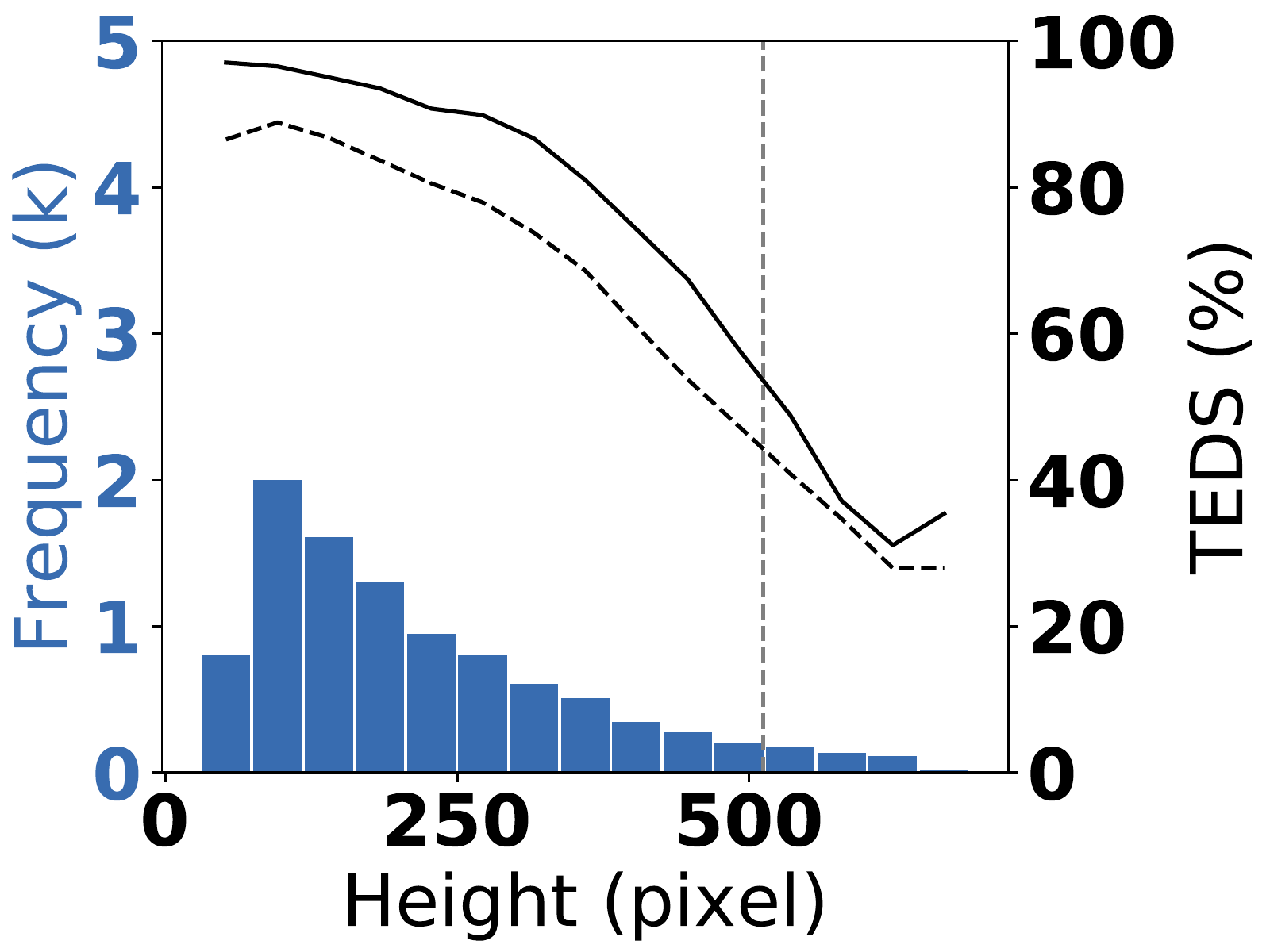}}%
  \end{minipage}\\
  \begin{minipage}[b]{.5\linewidth}%
  \subfloat{\includegraphics[width=\linewidth]{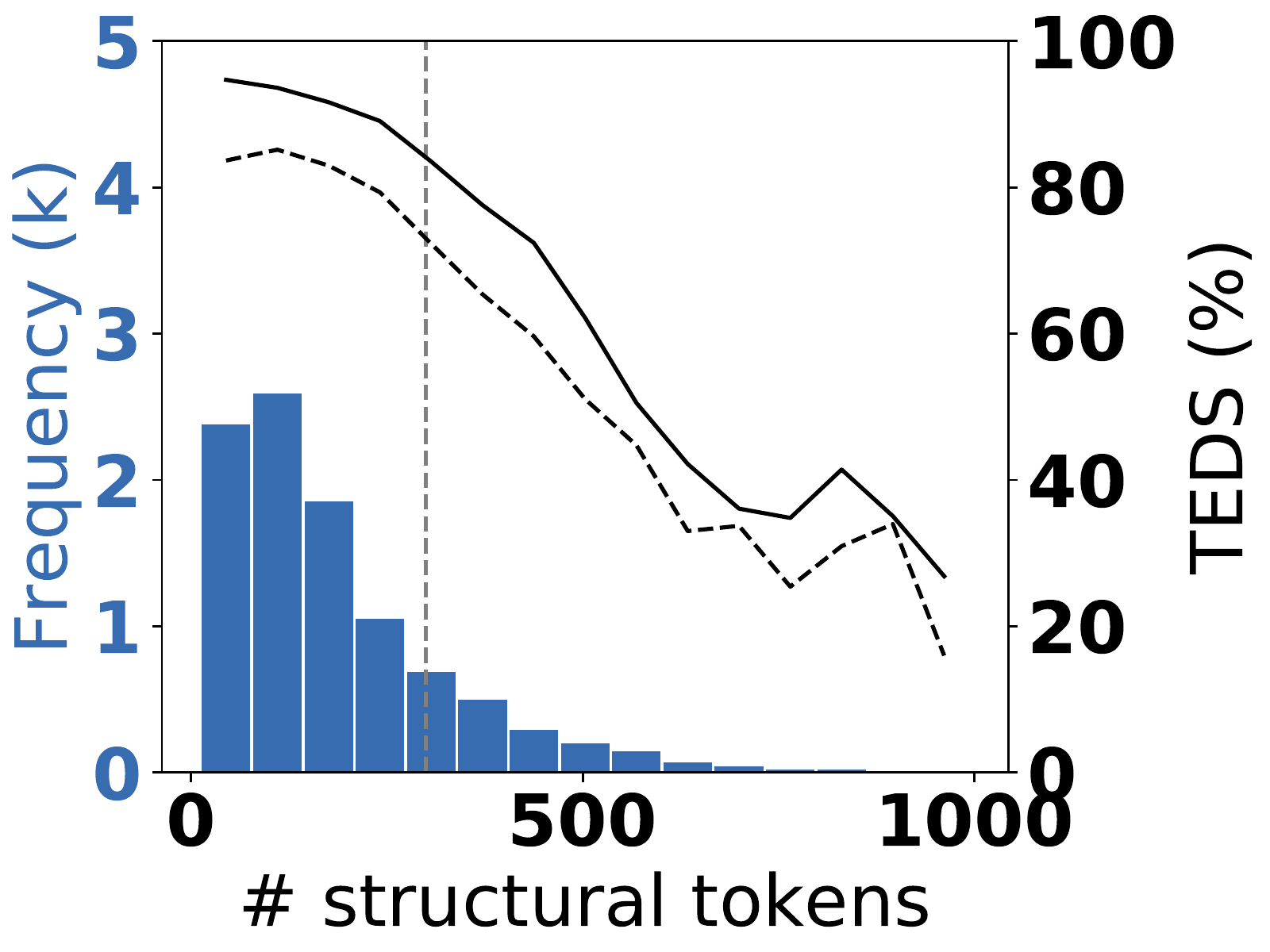}}%
  \end{minipage}%
  \begin{minipage}[b]{.5\linewidth}%
  \subfloat{\includegraphics[width=\linewidth]{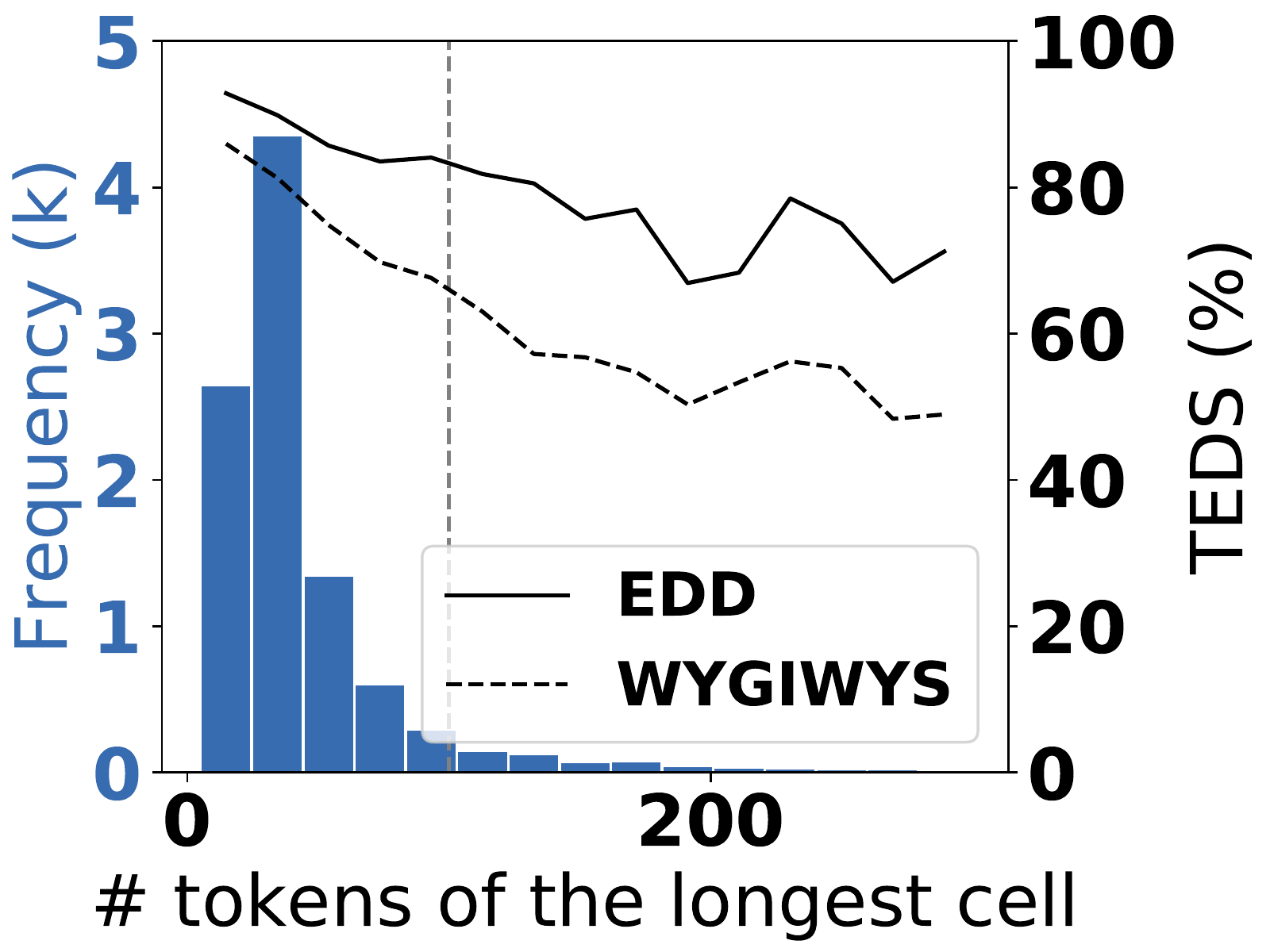}}%
  \end{minipage}
  \caption{Impact of table size in terms of width, height, number of structural
  tokens, and number of tokens in the longest cell on the performance of EDD and
  WYGIWYS. The bar plots (left axis) are the histogram of PubTabNet test set
  \wrt the above properties. The line plots (right axis) are the mean TEDS of
  the samples in each bar. The vertical dashed lines are the thresholds in
  Equation~\ref{eq:small}.}
  \label{fig:error_analysis}
\end{figure}

We categorize the test set of PubTabNet into 15 equal-interval groups along four
key properties of table size: width, height, number of structural tokens, and
number of tokens in the longest cell. Fig.~\ref{fig:error_analysis}
illustrates the number of tables in each group and the performance of the EDD
model and the WYGIWYS model on each group. The EDD model outperforms the WYGIWYS
model on all groups. The performance of both models decreases as table size
increases. We train the models with tables that satisfy Equation~\ref{eq:small},
where the thresholds are indicated with vertical dashed lines in
Fig.~\ref{fig:error_analysis}. Except for width, we do not observe a steep
decrease in performance near the thresholds. We think the lower performance on
larger tables is mainly due to rescaling images for batching, where larger
tables are more strongly downsampled. The EDD model may better handle large
tables by grouping table images into similar sizes as in~\cite{deng2017image}
and using different rescaling sizes for each group.

\subsection{Generalization}

\begin{figure}[!ht]
  \centering
  \begin{minipage}[b]{\linewidth}%
  \subfloat[Category 1]{\includegraphics[width=\linewidth]{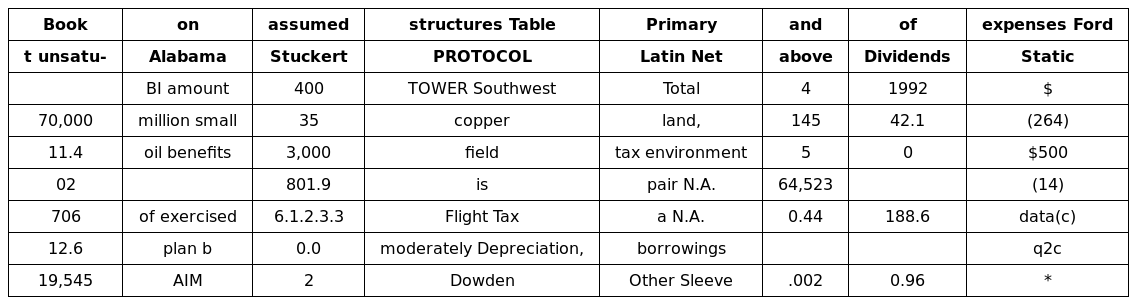}}%
  \end{minipage}\\
  \begin{minipage}[b]{\linewidth}%
  \subfloat[Category 2]{\includegraphics[width=\linewidth]{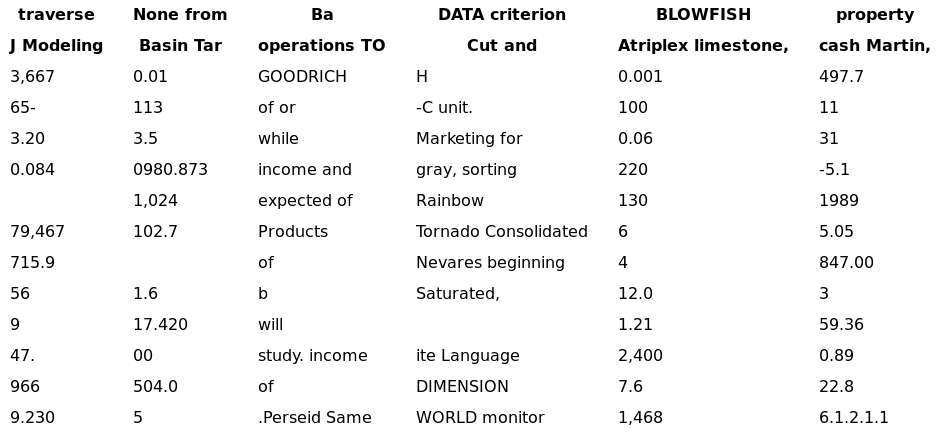}}%
  \end{minipage}\\
  \begin{minipage}[b]{\linewidth}%
  \subfloat[Category 3]{\includegraphics[width=\linewidth]{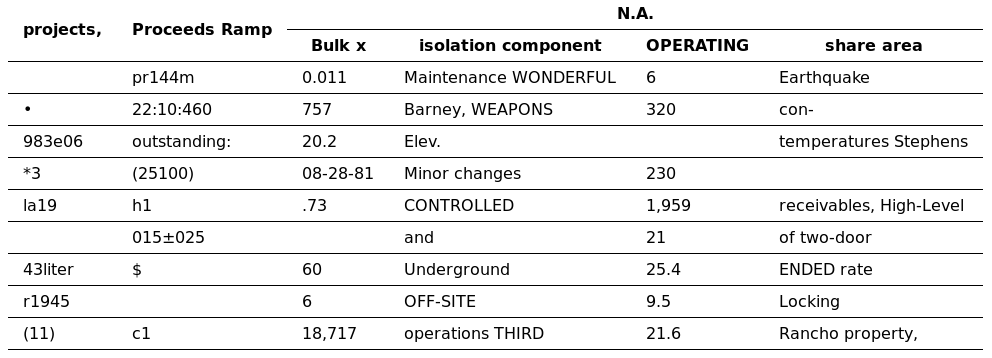}}%
  \end{minipage}\\
  \begin{minipage}[b]{\linewidth}%
  \subfloat[Category 4]{\includegraphics[width=\linewidth]{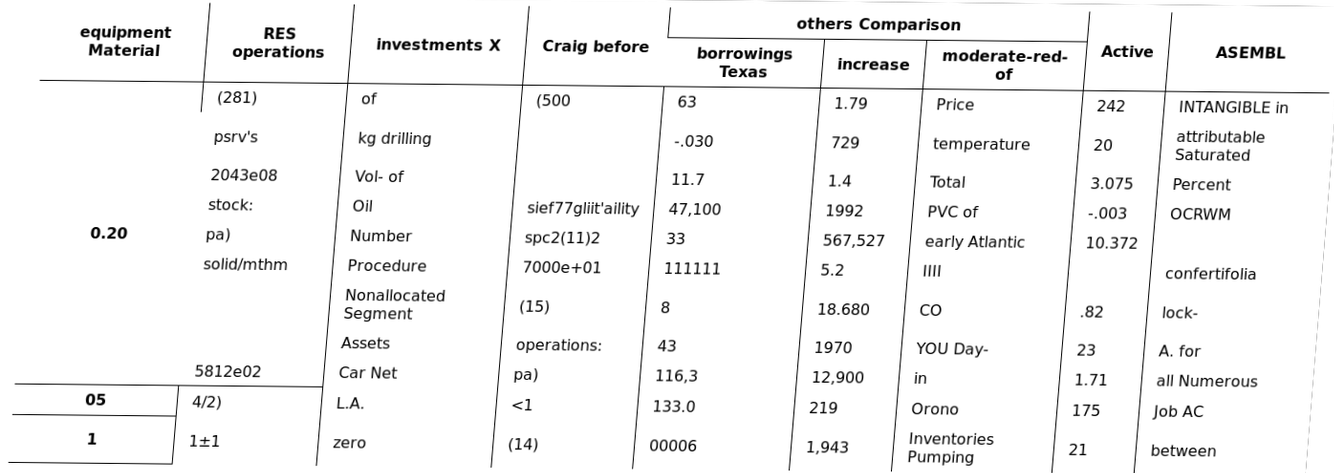}}%
  \end{minipage}
  \caption{Sample table image of the four categories of table styles defined in~\cite{qasim2019rethinking}.}
  \label{fig:style}
\end{figure}

To demonstrate that the EDD model is not only suitable for PubTabNet, but also
generalizable to other table recognition datasets, we train and test EDD on the
synthetic dataset proposed in~\cite{qasim2019rethinking}. We did not choose the
ICDAR2013 or ICDAR2019 table recognition competition datasets. Because, as shown
in Table~\ref{tab:dataset}, ICDAR2013 does not provide enough training data; and
ICDAR2019 does not provide ground truth of cell content (cell position only). We
synthesize 500K table images with the corresponding HTML
representation\footnote{\url{https://github.com/hassan-mahmood/TIES\_DataGeneration}},
evenly distributed among the four categories of table styles defined
in~\cite{qasim2019rethinking} (see Fig.~\ref{fig:style} for
example). The synthetic data is partitioned (stratified sampling by category)
into 420K/40k/40k training/validation/test sets.

We compare the test performance of EDD to the graph neural network model TIES
proposed in~\cite{qasim2019rethinking} on each table category. We compute the
TEDS score only for EDD, as TIES predicts if two tokens (recognized by an OCR
engine from the table image) share the same cell, row, and column, but not a
HTML representation of the table\footnote{\cite{qasim2019rethinking} does not
describe how the adjacency relations can be converted to a unique HTML
representation.}. Instead, as in~\cite{qasim2019rethinking}, the exact match
percentage is calculated and compared between EDD and TIES. Note that the exact
match for TIES only checks if the cell, row, and column adjacency matrices of
the tokens perfectly match the ground truth, but does not check if the OCR
engine makes any mistakes. For a fair comparison, we also ignore cell content
recognition errors when checking the exact match for EDD, i.e., the recognized
table is considered as an exact match as long as the structure perfectly matches
the ground truth.

Table~\ref{tab:test_TIES} shows the test performance of EDD and TIES, where EDD
achieves an extremely high TEDS score (99.7+\%) on all the categories of the
synthetic dataset. This means EDD is able to nearly perfectly reconstructed both
the structure and cell content from the table images. EDD outperforms TIES in
terms of exact match on all table categories. In addition, unlike TIES, EDD does
not show any significant downgrade in performance on category 3 or 4, in which
the samples have a more complex structure. This demonstrates that EDD is much
more robust and generalizable than TIES on more difficult examples.

\begin{table}[!htb]
  \begin{center}
    \begin{tabularx}{.85\linewidth}{lccccccccc}
      \toprule
      \multirow{2}{*}{Model} & \multicolumn{4}{c}{Average TEDS (\%)} & & \multicolumn{4}{c}{Exact match (\%)} \\
        \cline{2-5}\cline{7-10}
         & C1 & C2 & C3 & C4 & & C1 & C2 & C3 & C4 \\
      \midrule
      TIES & \--- & \--- & \--- & \--- & & 96.9 & 94.7 & 52.9 & 68.5 \\
      EDD & 99.8 & 99.8 & 99.8 & 99.7 & & 99.7 & 99.9 & 97.2 & 98.0 \\
      \bottomrule
    \end{tabularx}
  \end{center}
\caption{Test performance of EDD and TIES on the dataset proposed in~\cite{qasim2019rethinking}. TEDS score is not computed for TIES, as it does not generate the HTML representation of input image.}
\label{tab:test_TIES}
\end{table}

\section{Conclusion}

This paper makes a comprehensive study of the image-based table recognition
problem. A large-scale dataset PubTabNet is developed to train and evaluate deep
learning models. By separating table structure recognition and cell content
recognition tasks, we propose an attention-based EDD model. The structure
decoder not only recognizes the structure of input tables, but also helps the
cell decoder to place its attention on the right cell content. We also propose a
new evaluation metric TEDS, which captures both the performance of table
structure recognition and cell content recognition. Compare to the traditional
adjacency relation metric, TEDS can more appropriately capture multi-hop cell
misalignment and OCR errors. The proposed EDD model, when trained on PubTabNet,
is effective on recognizing complex table structures and extracting cell content
from image. PubTabNet has been made available and we believe that PubTabNet will
accelerate future development in table recognition and provide support for
pre-training table recognition models.

Our future works will focus on the following two directions. First, current
PubTabNet dataset does not provide coordinates of table cells, which we plan to
supplement in the next version. This will enable adding an additional branch to
the EDD network to also predict cell location. We think this additional task
will assist cell content recognition. In addition, when tables are available in
text-based PDF format, the cell location can be used to extract cell content
directly from PDF without using OCR, which might improve the overall recognition
quality. Second, the EDD model takes table images as input, which implicitly
assumes that the accurate location of tables in documents is given by users. We
will investigate how the EDD model can be integrated with table detection neural
networks to achieve end-to-end table detection and recognition.

\bibliographystyle{IEEEtran}
\bibliography{PubTabNet}

\end{document}